\documentclass[conference]{IEEEtran}
\usepackage{cite}

\ifCLASSINFOpdf
\else
\fi
\usepackage{multicol}

\usepackage{times}
\usepackage{latexsym}
\usepackage{algorithm}
\usepackage{algorithmic}
\usepackage{listings}
\usepackage[utf8]{inputenc}
\usepackage[T1]{fontenc}
\usepackage{amsmath}
\usepackage{amsfonts}
\usepackage{amssymb}
\usepackage[version=4]{mhchem}
\usepackage{stmaryrd}
\usepackage{float}

\usepackage{multirow}
\usepackage{enumitem}
\usepackage{booktabs}
\usepackage{rotating}
\usepackage{pgfplots}
\usepackage{rotating}
\usepackage{pgfplotstable}
\usepackage{tabularx}
\usepackage{graphicx} 
\usepackage{float} 
\usepackage{subfigure} 

\usepackage{latexsym}
\usepackage{graphicx}
\usepackage{url}

\usepackage{soul}

\usepackage{array}
\usepackage{xcolor}
\usepackage{adjustbox}
\usepackage{makecell}

\usepackage{caption}
\usepackage{graphicx}
\usepackage{amsmath}
\usepackage{booktabs}
\usepackage{algorithm}
\usepackage{algorithmic}
\usepackage{float}

\usepackage{newfloat}
\usepackage{listings}
\usepackage{algorithm}
\usepackage{algorithmic}
\usepackage[T1]{fontenc}

\usepackage[utf8]{inputenc}

\usepackage{microtype}

\usepackage{inconsolata}
\begin{document}
%
\title{Leveraging Professional Radiologists' Expertise to Enhance  LLMs' Evaluation for Radiology Reports}

\author{\IEEEauthorblockN{Qingqing Zhu\IEEEauthorrefmark{1}, Xiuying Chen\IEEEauthorrefmark{2}, Qiao Jin\IEEEauthorrefmark{1}, Benjamin Hou\IEEEauthorrefmark{3}, Tejas Sudharshan Mathai\IEEEauthorrefmark{3},\\
Pritam Mukherjee\IEEEauthorrefmark{3}, Xin Gao\IEEEauthorrefmark{2}, Ronald M. Summers\IEEEauthorrefmark{3}, Zhiyong Lu\IEEEauthorrefmark{1}}
\IEEEauthorblockA{\IEEEauthorrefmark{1}National Center for Biotechnology Information, National Library of Medicine,\\
National Institutes of Health, Bethesda, MD, USA\\ }
\IEEEauthorblockA{\IEEEauthorrefmark{2}Bioscience Reseach Center, King Abdullah University of Science \& Technology, Saudi Arabia\\}
\IEEEauthorblockA{\IEEEauthorrefmark{3}Imaging Biomarkers and Computer-Aided Diagnosis Laboratory,\\
Department of Radiology and Imaging Sciences,\\
National Institutes of Health Clinical Center, Bethesda, MD, USA\\ }}

\maketitle

\begin{abstract}
In radiology, Artificial Intelligence (AI) has significantly advanced report generation, but automatic evaluation of these AI-produced reports remains challenging. 
Current metrics, such as Conventional Natural Language Generation (NLG) and Clinical Efficacy (CE), often fall short in capturing the semantic intricacies of clinical contexts or overemphasize clinical details, undermining report clarity. 
To overcome these issues, our proposed method synergizes the expertise of professional radiologists with Large Language Models (LLMs), like GPT-3.5 and GPT-4. 
Utilizing In-Context Instruction Learning (ICIL) and Chain of Thought (CoT) reasoning, our approach aligns LLM evaluations with radiologist standards, enabling detailed comparisons between human and AI-generated reports. 
This is further enhanced by a Regression model that aggregates sentence evaluation scores.  
Experimental results show that our “Detailed GPT-4 (5-shot)” model achieves a 0.48 score, outperforming the METEOR metric by 0.19, 
while our “Regressed GPT-4” model shows even greater alignment with expert evaluations, exceeding the best existing metric by a 0.35 margin. 
Moreover, the robustness of our explanations has been validated through a thorough iterative strategy.  
We plan to publicly release annotations from radiology experts, setting a new standard for accuracy in future assessments. This underscores the potential of our approach in enhancing the quality assessment of AI-driven medical reports.

\end{abstract}


%
\IEEEpeerreviewmaketitle

\section{Introduction}

With the progression of Artificial Intelligence (AI) and Machine Learning (ML) technologies, automated report generation systems are increasingly leveraging a myriad of models \cite{zhu2021combining,chen2023topic,chen2023improving}. The accuracy and precision of generated outputs are paramount in medical fields, such as radiology, due to their direct impact on patient care \cite{wu2022multimodal,kaur2022chexprune,mohsan2022vision,jin2024hidden}. As such, establishing an effective and accurate evaluation framework for these reports is essential.

While radiologist assessments are considered the gold standard for evaluating radiology reports, relying on their expertise and context-specific knowledge, the growing volume of AI-generated reports makes this approach increasingly impractical. Current automatic evaluation metrics, including Natural Language Generation (NLG) and Clinical Efficacy (CE) metrics, offer efficiency but are often limited in capturing the depth and complexity necessary for medical reports \cite{irvin2019chexpert}. Furthermore, a significant shortcoming of these metrics is their lack of explanatory power. They typically provide scores without detailed explanations, leaving a gap in understanding the nuances and clinical relevance of the reports. This limitation underscores the need for more advanced evaluation tools that can provide not just quantitative assessments but also qualitative insights.

Recent advancements in the field of natural language processing, particularly with Large Language Models (LLMs), present potential solutions to these challenges.  Groundbreaking studies  \cite{devlin-etal-2019-bert, radford2019language, wei2023symbol, ye2023context, brown2020language, openai2023gpt4, chowdhery2022palm, anil2023palm} have demonstrated the adaptability of LLMs in diverse tasks. 
This adaptability is realized through explicit instructions and few-shot templates, a paradigm often referred as In-Context Instruction Learning (ICIL) \cite{ye2023context}. 
Additionally, the concept of Chain of Thought (CoT) reasoning, which involves a series of intermediate reasoning steps, significantly enhances the capability of LLMs for complex reasoning tasks, as indicated by \cite{wei2022chain}. This methodology, already successfully applied in various general domains, hints at its immense potential when applied to in-depth medical reasoning, particularly under the guidance of expert radiologists.
By harnessing these advancements, our method aims to combine the expertise of radiologists with the strengths of LLMs, creating a new way to evaluate complex medical reports.

 Our contributions are:
(1) Introducing a unique approach that combines radiologist expertise, ICIL, and CoT to improve the evaluation of radiology reports with LLMs, notably GPT-3.5 and GPT-4.
(2) Benchmarking our method against existing metrics, quantifying its correlation with expert evaluations, and demonstrating its superiority over the state-of-the-art. Additionally, we plan to publicly release annotations from radiology experts, setting a new standard for accuracy in future assessments.
(3) Our approach not only offers a comprehensive and precise assessment of AI-generated radiology reports but also provides the added benefit of explainability.

\section{Related Work}
\subsection{Evaluation Metrics in Radiology Reports}

Several metrics have been developed and used for evaluating text generated by AI systems. Metrics like BLEU (Bilingual Evaluation Understudy) \cite{papineni2002bleu}, METEOR (Metric for Evaluation of Translation with Explicit ORdering) \cite{banerjee2005meteor}, and ROUGE (Recall-Oriented Understudy for Gisting Evaluation) \cite{lin2004rouge} are used widely, but they each have limitations when applied to medical reports \cite{novikova2017we,pang2023survey}. These metrics primarily assess the n-gram overlap between generated text and reference text, or consider word and phrase alignments. Consequently, they lack the capacity to evaluate complex semantic and contextual nuances that are intrinsic to medical reporting \cite{kilickaya2016re}.
Meanwhile, CE metrics such as F1 score, precision, and recall, primarily used in machine learning, have been adapted to evaluate the performance of automated systems in identifying and categorizing observations in radiology reports \cite{irvin2019chexpert,sokolova2009systematic,chen2018deep}. While these metrics are proficient in assessing the model's ability to correctly identify observations, they fall short in evaluating the overall quality and coherence of the generated reports.

In contrast, our proposed method provides a more nuanced evaluation of AI-generated radiology reports. It adeptly captures the essential details and subtleties inherent in such reports, making it particularly applicable in the medical domain. Furthermore, our approach not only saves substantial human labor by reducing the reliance on manual evaluation but also possesses the unique quality of being explainable, further enhancing its practical value.

\subsection{LLMs for Evaluation}
In recent years, we've seen significant advancements in LLMs, with models spanning from BERT \cite{DBLP:conf/naacl/DevlinCLT19} to GPT. These models, characterized by their escalating sophistication and capabilities, have vastly facilitated the progression of advanced techniques. Among them, GPT-3.5 and GPT-4 have catalyzed a paradigm shift within the field of intelligent human-machine dialogue. This shift continues to make a significant impact on the research community and various industries \cite{tian2023opportunities}.

The advent of ChatGPT has stimulated immense interest in two primary areas. Firstly, many papers explore its performance across a myriad of Natural Language Processing (NLP) tasks, shedding light on its extensive capabilities. Secondly, there's growing intrigue in employing it as a metric for evaluating model outputs \cite{DBLP:journals/corr/abs-2302-14520}. Evaluations involving ChatGPT typically fall into two categories: Natural Language Understanding (NLU) and Natural Language Generation (NLG). ChatGPT has demonstrated remarkable performance across virtually all NLU tasks, as confirmed by existing work \cite{DBLP:journals/corr/abs-2302-06476,DBLP:journals/corr/abs-2302-04023}. Within the NLG sphere, it has been applied in areas such as machine translation \cite{DBLP:journals/corr/abs-2304-02426}, monolingual summarization \cite{gao2023umse}, cross-lingual summarization \cite{DBLP:journals/corr/abs-2302-14229}, review generation \cite{DBLP:conf/inlg/WangZHKJR20}, and radiology reports generation \cite{DBLP:journals/corr/abs-2212-14882}. However, our work diverges by employing GPT-3.5 or GPT-4 as a human evaluator, using it to autonomously assess the quality of general textual generations, rather than merely utilizing it to solve tasks.

While studies exist that utilize ChatGPT to evaluate specific fields such as translation \cite{DBLP:journals/corr/abs-2302-14520} or human personalities \cite{DBLP:journals/corr/abs-2303-01248}, these applications are often simplistic and lack grounding in any professional domain. In contrast, our study innovatively employs GPT-3.5 or GPT-4 in the medical domain, which also actively involves domain experts in the process.

\section{Method}

Our research is uniquely positioned to evaluate any model designed for generating radiological reports, showcasing a broad applicability in this field. 

In our study, we primarily concentrate on the model presented in Zhu \cite{zhu2023utilizing}, which we refer to as the LongiFill model in our paper. This choice is motivated by its recent advancements in the field of medical report generation.
This model leverages longitudinal multi-modal data, encompassing prior patient visit chest X-rays (CXR), current visit CXR, and the previous visit's report, to efficiently pre-populate the report for a current patient visit. We conduct our evaluation using reports generated by the this model that is trained on the MIMIC-CXR \footnote{https://physionet.org/content/mimic-cxr-jpg/2.0.0/}  dataset. 
It's important to note that our evaluation framework is distinct from the LongiFill model. 
In report generation process, we treat reports from the MIMIC-CXR dataset as ``Original'' and those generated by AI models as ``Predicted'' reports. This framework integrates the expertise of professional radiologists with advanced LLMs, specifically GPT-3.5 and GPT-4.This integration enables a more comprehensive and nuanced analysis of the generated reports.

Figure \ref{overall} shows the whole architecture of our evaluation strategy, following the steps:



\begin{figure*}[!h]
\centering
\includegraphics[width=0.8\textwidth]{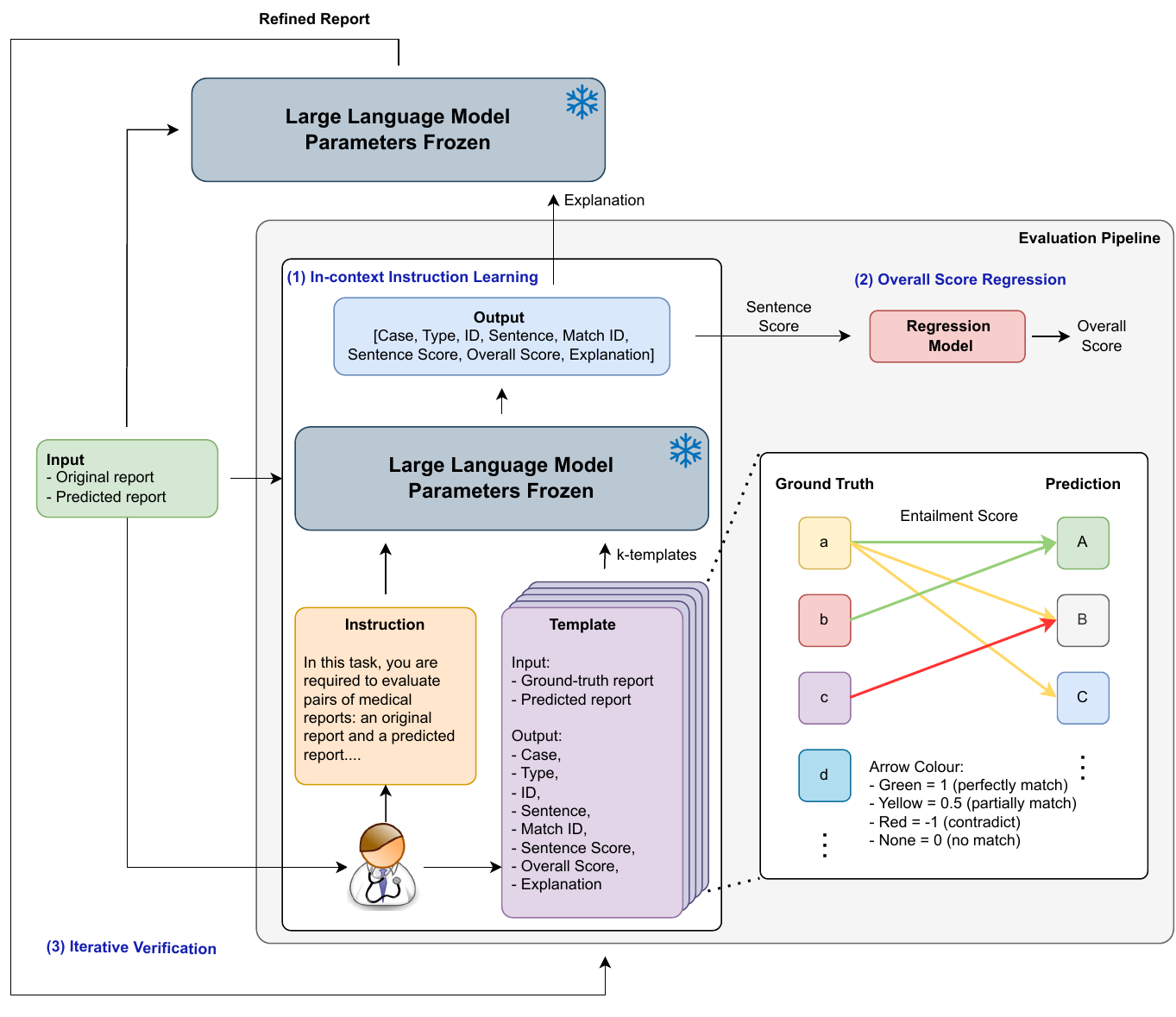}
\caption{The whole architecture of our evaluation strategy. It is primarily focused on three key areas: In-context Instruction Learning, Overall Score Regression and Iterative Verification. The ``sentence score'' within the template represents the entailment score, derived by comparing each sentence from the original reports with its corresponding sentence in the prediction. An explanation for this is provided in the lower right corner of the figure.}

\label{overall}
\end{figure*}


(1) First, the input, which includes the Original and Predicted reports, is passed to an LLM, along with instructions and templates. The number of these templates is denoted as ``k''. This stage epitomizes our ICIL approach. It ensures that the model not only determines scores but also formulates an explanation with a set output structure, following a logical sequence. The ``Explanation'' segment, though not delineated into a sequential chain of thought, mirrors the essence of a CoT. It offers clarity and sheds light on the derivation of the overall score. Grounded in the CoT methodology, this explanation acts as a pivotal intermediate output, setting the foundation for further enhancements.
(2) The sentence scores derived from the ICIL process are then fed into a regression model. The objective is to calculate an overall score for the report. This phase is labeled ``Overall Score Regression''.
(3) In this phase, we focus on verifying the quality of our explanations. These explanations, along with the AI generated report, are then passed to the LLMs again. The result is a Refined Report. Subsequently, this report is returned for another round of evaluation.  This step is termed ``Iterative Verification'' and acts as our validation loop.

\subsection{In-context Instruction Learning}



\noindent\textbf{Large Language Models}  
For evaluation, we use LLMs such as GPT-3.5 and GPT-4. These models not only serve as quasi-human evaluators but also follow ICIL and CoT mechanisms, progressively building upon prior information to generate comprehensive outputs.

\noindent\textbf{Input} We first pre-process the radiology reports by splitting them using a period (``.''). Following this segmentation, each sentence is then identified using specific identifiers such as ``a'', ``b'', and so on, in a sequential manner. 

\noindent\textbf{Role of Radiologists} Radiologists play a crucial role in crafting the \textbf{instructions}, designing the evaluation \textbf{templates}, and labeling  the cases. These instructions and templates were incorporated into the LLMs, which subsequently assisted the LLMs in emulating professional evaluation patterns. The labeled cases were subsequently utilized for in-depth analysis.

\noindent\textbf{Instruction}
Two types of instructions were utilized: the simplistic and detailed versions (We outline both versions of the task instructions in the supplementary materials). Both sets of instructions were crafted with the assistance of a renowned radiology expert. The simplistic version is concise, directing radiologists to compare predictions with the original reports and then provide an overall score. Conversely, the detailed version elaborates on the nature of the reports, introducing a well-defined scoring system, and underscoring the real-world significance of the task. 
These instructions guide LLMs through the evaluation process.

\noindent\textbf{Templates}
Each template consists of two distinct sections: \textbf{input} and \textbf{output}. A representative example of this can be found in the supplementary materials. The structure for the \textbf{input} adheres to the previously specified format. The \textbf{output} section, however, is delineated into the subsequent fields:
\textbf{1. Case}: Denotes the number assigned to each case. 
 \textbf{2. Type}: Distinguishes whether the sentence is sourced from the original reports or is a prediction.
\textbf{3. ID}: A unique identifier assigned to each sentence.
\textbf{4. Sentence}: The actual content of the sentence.
\textbf{5. Match ID}: An identifier linking to matched sentences.
 \textbf{6. Sentence Score}: A numerical value that signifies the level of entailment between the predicted and original statements. The scoring metrics are:
(1) Score of 1: The predicted statement aligns in meaning and detail with the original. Word choice may vary, but the core message remains consistent. Crucially, no original information is omitted or contradicted.
(2) Score of 0.5: The predicted statement bears some resemblance to the original statement, though not entirely. While certain elements are congruent, others may differ or be absent. The foundational idea is present, but not every detail is precise.
(3) Score of -1: The predicted statement is diametrically opposed to or starkly contrasts the original statement. Any information provided directly challenges the original content.
(4) Score of 0: The predicted statement lacks a corresponding original statement for comparison.
7. In addition to the detailed output for each sentence, every case comes with:
 \textbf{(1) Overall Score}: Represents a holistic assessment of the example from 0 to 5. 
\textbf{(2) Explanation}: A descriptive commentary elucidating the CoT that provides context and insight into how the overall score was derived.

\subsection{Overall Score Regression}
Instead of solely relying on the overall scores directly assigned by GPT-3.5 or GPT-4, we adopted an alternative approach that employed regression models to predict the overall score.  By leveraging this model, our goal was to encapsulate the combined insights of multiple features,  and potentially gain a more nuanced understanding of the underlying patterns in the data \cite{segal2004machine}.

\textbf{1. Feature Engineering: Sentence Score Ratios}To establish the input features for our regression model, we calculated the ratios of various sentence scores in both original and AI-predicted sentences. These scores, representing key assessment criteria, include 0, 0.5, 1, and -1. The ratios, indicating the frequency of each score relative to the total, are crucial in understanding the balance of scoring tendencies. For the original report sentences, with a total count of \( m \), the ratios for scores 0, 0.5, 1, and -1 are denoted as \( r_{o_0} \), \( r_{o_{0.5}} \), \( r_{o_1} \), and \( r_{o_{-1}} \) respectively. Similarly, for the AI-predicted sentences having a total count of \( n \), the ratios are \( r_{p_0} \), \( r_{p_{0.5}} \), \( r_{p_1} \), and \( r_{p_{-1}} \). The feature set \(\mathbf{X}\) is thus represented as:
\[
\mathbf{X} = [r_{o_0}, r_{o_{0.5}}, r_{o_1}, r_{o_{-1}}, r_{p_0}, r_{p_{0.5}}, r_{p_1}, r_{p_{-1}}]
\]

\textbf{2. Target Variable}
The overall scores given by human annotators, denoted by \( O \), serve as our target variable \(\mathbf{Y}\), which represent aggregate or average evaluations of specific criteria by annotators, encapsulating a holistic view of report quality.

\textbf{3. Training }
With the defined feature set \(\mathbf{X}\) and target variable \(\mathbf{Y}\), we employed various regression models (e.g., Decision Tree, Support Vector Machine, K-Nearest Neighbors, Neural Network, Gradient Boosting, Random Forest) to assess the performance of LLMs. Each model predicts regressed overall scores (\(\hat{\mathbf{Y}}\)) based on the given features and is represented as:
\[
\hat{\mathbf{Y}} = f_{\text{model}}(\mathbf{X}; \boldsymbol{\theta})
\]
where \( f_{\text{model}} \) denotes the specific regression model used, and \( \boldsymbol{\theta} \) represents the model parameters.

\subsection{Iterative Verification of the Explanatory Mechanisms}

Large Language Models (LLMs) do more than just evaluate AI reports. They offer detailed insights into AI decisions, showing both their strengths and weaknesses. To validate the accuracy of these explanations, we first employ a refined process. This process is structured around specific instructions and templates (examples of this process can be found in the supplementary materials). These instructions carefully steer the creation of polished reports based on the explanations provided, while the template serves as a practical example, further clarifying the refinement process. The Refined Report then goes through another round of evaluation. No additional data, such as the original image, was introduced during this process. Then by re-evaluating the Refined Report, it's possible to assess whether the initial explanations were accurate and whether they have been effectively integrated into the new version of the report.

\section{Experiments}

We utilized both GPT-4 and GPT-3.5. For our work, we employed the OpenAI API version ``2023-03-15-preview'', using the engines ``gpt-35-turbo'' for GPT-3.5 and ``gpt-4'' for GPT-4. We find that the scores predicted by the models for each sentence could vary slightly due to the nature of sampling.  Therefore, to ensure accuracy and reliability in our findings, we conducted three iterations of the sentence scoring process, observing some minor variations and calculated average results for the training data used in our regressed model and evaluation process.

\noindent \textbf{Metrics for Comparison}
 We juxtaposed our proposed method against several established automatic evaluation metrics:
    \textbf{1. NLG Metrics:} Metrics such as BLEU \cite{papineni2002bleu}, METEOR \cite{banerjee2005meteor}, and \(\text{Rouge}_L\) \cite{lin2004rouge} were incorporated into our comparative study.
    \textbf{2. CE Metrics:} For clinical accuracy assessment of generated reports, we employed the CheXpert labeler\footnote{\url{https://github.com/stanfordmlgroup/chexpert-labeler/}} \cite{irvin2019chexpert}. Emulating the methodology of \cite{moon2022multi}, we contrasted the positive labels of 14 CheXpert observations\footnote{No Finding, Enlarged Cardiomediastinum, Cardiomegaly, Lung Lesion, Airspace Opacity, Edema, Consolidation, Pneumonia, Atelectasis, Pneumothorax, Pleural Effusion, Pleural Other, Fracture, and Support Devices} based on accuracy, precision, recall, and F-1 metrics.

\noindent\textbf{Human Assessment}
The human assessment involved 100 original-prediction pairs, which were randomly selected by the LongiFill model. It was carried out by three human raters: Rater1 is a physician with a doctoral degree, while Rater2 and Rater3 were trained in biomedical informatics. A consensus, achieved after deliberation among the three raters, served as the \textbf{ground truth}. During this process, to guarantee the quality of annotations,  we also sought assistance from a renowned radiology expert. We carefully selected five of these cases as templates, which showcased diverse scoring patterns. These remaining 95 examples are used for further evaluation.

\section{Results and Analyses}\label{sec:Introduction}

\begin{figure}[h]
\centering

\begin{tikzpicture} [scale=0.40]
    \begin{axis}[
            x=1cm,
            y=1cm,
            colormap={bluewhite}{color=(white) rgb255=(90,96,191)},
            xlabel style={yshift=-30pt},
            ylabel style={yshift=20pt},
            xticklabels={bleu,meteor, rouge\_l,precision,recall,accuracy,f1\_score,  detailed Gpt-4 (5-shot), Regressed Gpt-4,detailed Gpt-3.5 (5-shot),ground truth}, 
            xtick={0,...,10},
            xtick style={draw=none},
            yticklabels={bleu,meteor, rouge\_l,precision,recall,accuracy,f1\_score, detailed Gpt-4 (5-shot),Regressed Gpt-4 (5-shot),detailed Gpt3.5 (5-shot),ground truth }, 
            ytick={0,...,10},
            ytick style={draw=none},
            enlargelimits=false,
            colorbar,
            xticklabel style={
              rotate=30,anchor=north east
            },
            yticklabel style={
              rotate=30,anchor=north east,yshift=12pt
            },
            nodes near coords={ \pgfmathfloatifflags{\pgfplotspointmeta}{0}{}{\pgfmathprintnumber[fixed]{\pgfplotspointmeta}}
},
            nodes near coords style={
                yshift=-7pt
            },
        ]
        \addplot[
            matrix plot,
            mesh/cols=11,
            point meta=explicit,draw=gray
        ] table [meta=C] {
            x  y C
0 0 0
1 0 0
2 0 0
3 0 0
4 0 0
5 0 0
6 0 0
7 0 0
8 0 0
9 0 0
10 0 0

0 1  0.5385858585858586
1 1 0
2 1 0
3 1 0
4 1 0
5 1 0
6 1 0
7 1 0
8 1 0
9 1 0
10 1 0

0 2 0.6028282828282829
1 2 0.5397979797979799
2 2 0
3 2 0
4 2 0
5 2 0
6 2 0
7 2 0
8 2 0
9 2 0
10 2 0

0 3 0.1198
1 3 0.0827
2 3 0.0793
3 3 0
4 3 0
5 3 0
6 3 0
7 3 0
8 3 0
9 3 0
10 3 0

0 4 0.0553
1 4 0.1191
2 4 0.0452
3 4 0.2813
4 4 0
5 4 0
6 4 0
7 4 0
8 4 0
9 4 0
10 4 0

0 5 0.0753483629659654
1 5 0.07345874884769667
2 5 0.042752519425829895
3 5 0.7053171338739675
4 5 0.682112805501613
5 5 0
6 5 0
7 5 0
8 5 0
9 5 0
10 5 0

0 6 0.23151490100773875
1 6 0.2876002009701769
2 6 0.26760079168124545
3 6 0.07260436934897734
4 6 0.17160331887053606
5 6 0.11793308479883252
6 6 0
7 6 0
8 6 0
9 6 0
10 6 0

0 7 0.21590503204589445
1 7 0.2342165041917987
2 7 0.28489243873511516
3 7 0.0737813047906248
4 7  0.14871359154945354
5 7 0.1366738697057346
6 7 0.1326023405543814
7 7 0
8 7 0
9 7 0
10 7 0

0 8 0.1371855750004308
1 8 0.19456510857940332
2 8 0.16304677323320715
3 8 0.07022089254020247
4 8 0.19455737327138006
5 8 0.17244209027863838
6 8 0.14055424816215342
7 8 0.45336119403155856
8 8 0
9 8 0
10 8 0

0 9  0.14222574607030908
1 9 0.08083016003374709
2 9 0.10512337753022845
3 9 0.007837087730606604
4 9 0.01
5 9 0.01607966834025456
6 9 0.010616043579314852
7 9 0.19086935856665976
8 9 0.14334446319929559 
9 9 0
10 9 0

0 10 0.23151490100773875
1 10 0.2876002009701769
2 10 0.26760079168124545
3 10 0.07260436934897734
4 10 0.17160331887053606
5 10 0.11793308479883252
6 10 0.1551185116626245
7 10  0.47748206564958845
8 10 0.637
9 10  0.15518154608614856
10 10 0

        };
    \end{axis}

\end{tikzpicture}
\caption{Correlation matrix of Kendall's Tau Values for Metric Pairs. All scores have p value < 0.05.}
    \label{heatmap1}
\end{figure}
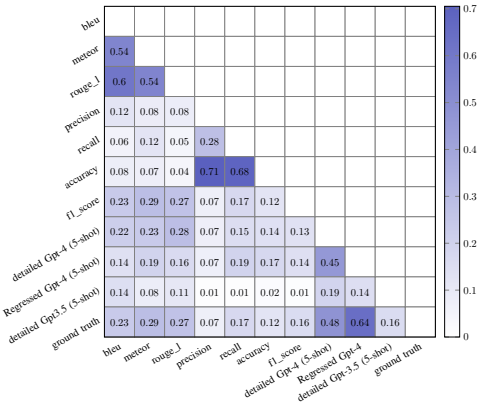

\noindent\textbf{Evaluating Alignment between Different Metrics and Human Evaluations} 
To discern the degree of alignment between automated metrics (including our methods) and human evaluations’ \textbf{overall score}, we relied on Kendall’s Tau \cite{kendall1938new}. 
Figure \ref{heatmap1} illustrates the results for various metric pairs.
\textbf{1. Comparative Metric Performance with our ``Detailed Gpt-4 (5-shot)'' Model} 
Among NLG metrics, METEOR demonstrates the strongest alignment score of 0.29 with the ground truth evaluations, followed by ROUGE and then BLEU. For the CE metrics, Recall and F1 score present the most significant correlation with the expert evaluations.
While NLG metrics generally outperform CE metrics in this context, both categories fall short when compared to the ``Detailed Gpt-4 (5-shot)''. This superior alignment may stem from the model's capability to discern nuanced and qualitative elements that aren't comprehensively captured by any single metric.
\textbf{2. Comparative Metric Performance with our ``Regressed Gpt-4'' Model}  We also evaluated the performance of the Regressed model. During  training, we used the ground truth sentence scores and overall scores from the 95 manually annotated examples. While we evaluated various regression models, the Random Forest model demonstrated the
best performance. Details of this ablation study can be found
in the supplementary materials. Subsequently, we input the sentence score from detailed GPT-4 (5-shot) into the Regressed model to obtain Regressed overall scores, denoted as "Regressed GPT-4" in the figure. This score was then compared with other evaluation methods. Notably, the Regressed model exhibits a correlation that is 0.35 (0.64 vs. 0.29) higher than the best METEOR in other metrics. This underscores the effectiveness of integrating machine learning with LLM evaluations.
\textbf{3. Comparative Metric Performance with our ``Detailed GPT-3.5 (5-shot)'' Model}  The performance of ``Detailed GPT-3.5 (5-shot)'' seems to lag behind both traditional metrics and GPT-4. This suggests that GPT-3.5 may struggle to accurately encapsulate the nuances within sentences.

\noindent\textbf{Iterative Verification Results}
A defining feature of our methodology is validation of the explanations produced. We randomly selected 100 reports from the LongiFill model. These chosen reports and their explanations were then fed into the GPT-4 model. After refinement, the reports underwent another round of evaluation. The  results of these evaluations are presented in Table \ref{rewrite}.
The table emphasizes notable improvements in the quality of the regenerated reports across all metrics, where the explanations has indeed assisted in producing superior reports. This iterative process illustrates the reliability of the explanations provided by our methodology and proves the interpretability of our framework.

\begin{table*}[ht!]\small
\begin{tabular}{l|l|p{7cm}|cc|cc} 
\hline
\textbf{Type} & \textbf{ID} & \textbf{Report}& \multicolumn{2}{c|}{Match ID}& \multicolumn{2}{c}{Score} 
\\ \cline{4-7}
 &  & & GPT-4 & GT &GPT-4 &GT\\ \hline
Original& a & Comparison is made to prior study from. &- & -&  0 & 0 \\ 
Original&b&There is a swan-ganz catheter whose distal lead tip is in the main pulmonary outflow tract. & -& - & 0  & 0 \\ 
Original & c & The cardiac silhouette is enlarged. & C & C& 1 &0.5 \\ 
Original& d & There is again seen moderate <unk> pleural effusion which is stable. & F& F &0.5& 0.5 \\ 
Original & e & There is some improvement in the pulmonary vascular edema. & E& E & 0.5& 0.5  \\ 
Original& f & There are no pneumothoraces identified. & H & H& 1 & 1\\ \hline
Prediction & A & The patient is status post median sternotomy and cabg. & - &- & 0 & 0 \\ 
Prediction & B & Left-sided AICD device is noted with leads terminating in the right atrium, right ventricle, and region of the coronary sinus.& - &- & 0 & 0\\ 
Prediction & C & Moderate to severe cardiomegaly is re-demonstrated. & c& c & 1& 0.5\\ 
 Prediction & D & The mediastinal contour is unchanged. & - & - & 0  & 0\\ 
Prediction & E & There is mild pulmonary vascular congestion. & e & e  & 0.5  & 0.5 \\ 
Prediction & F & Small bilateral pleural effusions are noted. & d & d& 0.5& 0.5 \\ 
Prediction & G & Patchy opacities in the lung bases likely reflect areas of atelectasis. & -& - & 0 & 0\\ 
Prediction & H & No pneumothorax is identified. & f & f& 1& 1 \\ \hline

\multicolumn{5}{l|}{Overall Score:} & 3 & 2.5 \\ \hline

 \multicolumn{7}{p{15cm}}{\textbf{Explanation:} The AI's generated report has some alignment with the ground truth - the cardiac silhouette enlargement and the absence of pneumothorax are accurate. The ground truth's note on improved pulmonary vascular edema aligns partially with the AI's note on mild congestion. However, the AI report doesn't highlight the presence of the swan-ganz catheter, and it inaccurately reports small bilateral effusion instead of the ground truth's moderate, stable pleural effusion. Additionally, the AI mentions post median sternotomy and cabg status, and AICD device installation, which are not mentioned in the ground truth.}
\\ \hline

\end{tabular}
\caption{Case study for evaluation of original and prediction reports. ``GT'' means ``Ground Truth''. }
\label{case}
\end{table*}
\begin{table*}[!ht]\small
\centering

\setlength\tabcolsep{0.9mm} 
\begin{tabular}{l|ccc|cccc|c}
\hline
& \multicolumn{3}{c|}{NLG} & \multicolumn{4}{c|}{CE} &         Our Method       \\ \cline{2-9} 
          & BL-1      & M      & $R_L$ & A    & P   & R   & F1   &  ReGPT-4\\ \hline
Before &    0.3060
&  0.1192
& 0.2389 & 0.7986&  0.4527& 0.346& 0.3922&2.4219
   \\
After  & \textbf{0.3846}
& \textbf{0.2420}
& \textbf{0.2480} & \textbf{0.8007}&\textbf{0.4716}&\textbf{0.5057}& \textbf{0.4881}&\textbf{3.5467}

\\ \hline
\end{tabular}
\caption{Comparative Analysis of Key Metrics ``Before'' and ``After'' Reintroducing Generated Reports into GPT-4. This iterative verification process aims to ensure the  reliability and correctness of explanations. }
\label{rewrite}
\end{table*}


\section{Discussion}
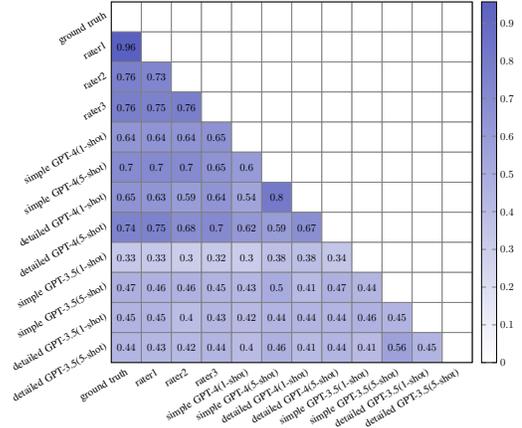
\begin{figure}[h]
\centering

\begin{tikzpicture} [scale=0.40]
    \begin{axis}[
            x=1cm,
            y=1cm,
            colormap={bluewhite}{color=(white) rgb255=(90,96,191)},
            xlabel style={yshift=-30pt},
            ylabel style={yshift=20pt},
            xticklabels={ground truth, rater1, rater2, rater3, simple GPT-4(1-shot), simple GPT-4(5-shot), detailed GPT-4(1-shot), detailed GPT-4(5-shot), simple GPT-3.5(1-shot), simple GPT-3.5(5-shot), detailed GPT-3.5(1-shot), detailed GPT-3.5(5-shot)}, 
            xtick={0,...,11}, 
            xtick style={draw=none},
            yticklabels={ground truth, rater1, rater2, rater3, simple GPT-4(1-shot), simple GPT-4(5-shot), detailed GPT-4(1-shot), detailed GPT-4(5-shot), simple GPT-3.5(1-shot), simple GPT-3.5(5-shot), detailed GPT-3.5(1-shot), detailed GPT-3.5(5-shot)}, 
            ytick={0,...,11}, 
            ytick style={draw=none},
            enlargelimits=false,
            colorbar,
            xticklabel style={
              rotate=30,anchor=north east
            },
            yticklabel style={
              rotate=30,anchor=north east,yshift=12pt
            },
            nodes near coords={  \pgfmathfloatifflags{\pgfplotspointmeta}{0}{}{\pgfmathprintnumber{\pgfplotspointmeta}}},
            nodes near coords style={
                yshift=-7pt
            },
        ]
        \addplot[
            matrix plot,
            mesh/cols=12, 
            point meta=explicit,draw=gray
        ] table [meta=C] {
            x  y C
            0  0 0
            1  0 0
            2  0 0
            3  0 0
            4  0 0
            5  0 0
            6  0 0
            7  0 0
            8  0 0
            9  0 0
            10 0 0
            11 0 0
            
            0  1 0.956
            1  1 0
            2  1 0
            3  1 0
            4  1 0
            5  1 0
            6  1 0
            7  1 0
            8  1 0
            9  1 0
            10 1 0
            11 1 0  
            
            0  2 0.757
            1  2 0.729
            2  2 0
            3  2 0
            4  2 0
            5  2 0
            6  2 0
            7  2 0
            8  2 0
            9  2 0
            10 2 0
            11 2 0 
    
            0  3 0.764
            1  3 0.746
            2  3 0.764
            3  3 0
            4  3 0
            5  3 0
            6  3 0
            7  3 0
            8  3 0
            9  3 0
            10 3 0
            11 3 0
    
            0  4 0.635
            1  4 0.639
            2  4 0.638
            3  4 0.648
            4  4 0
            5  4 0
            6  4 0
            7  4 0
            8  4 0
            9  4 0
            10 4 0
            11 4 0

            0  5 0.6974210147693142
            1  5 0.702
            2  5 0.697
            3  5 0.647
            4  5 0.602
            5  5 0
            6  5 0
            7  5 0
            8  5 0
            9  5 0
            10 5 0
            11 5 0

            0  6 0.646
            1  6 0.631
            2  6 0.587
            3  6 0.637
            4  6 0.543
            5  6 0.797322010287102
            6  6 0
            7  6 0
            8  6 0
            9  6 0
            10 6 0
            11 6 0

            0  7 0.743
            1  7 0.748
            2  7 0.675
            3  7 0.703
            4  7 0.617
            5  7 0.590
            6  7 0.667
            7  7 0
            8  7 0
            9  7 0
            10 7 0
            11 7 0

            0  8 0.331
            1  8 0.326
            2  8 0.300
            3  8 0.320
            4  8 0.304
            5  8 0.380
            6  8 0.383
            7  8 0.342
            8  8 0
            9  8 0
            10 8 0
            11 8 0

            0  9 0.469
            1  9 0.458
            2  9 0.458
            3  9 0.449
            4  9 0.429
            5  9 0.501
            6  9 0.412
            7  9 0.468
            8  9 0.438
            9  9 0
            10 9 0
            11 9 0

            0  10 0.447
            1  10 0.449
            2  10 0.397
            3  10 0.433
            4  10 0.415
            5  10 0.441
            6  10 0.443
            7  10 0.443
            8  10 0.459
            9  10 0.448
            10 10 0
            11 10 0

            0  11 0.442
            1  11 0.429
            2  11 0.420
            3  11 0.444
            4  11 0.404
            5  11 0.464
            6  11 0.412
            7  11 0.441
            8  11 0.409
            9  11 0.560
            10 11 0.450
            11 11 0

        }; 
    \end{axis}

\end{tikzpicture}

\caption{Correlation matrix depicting Cohen's Kappa scores for different annotation methods when aggregating sentence scores.} 
\label{heatmap}
\end{figure}

\begin{table*}[!ht]\small
\centering
\setlength\tabcolsep{0.9mm} 

\begin{tabular}{l|ccc|cccc|c}
\hline
& \multicolumn{3}{c|}{NLG} & \multicolumn{4}{c|}{CE} &         Our Method       \\ \cline{2-9} 
          & BL-1      & M      & $R_L$ & A    & P   & R   & F1   &  ReGPT-4
        \\ \hline
          
Transformer
&  0.2951
 &0.1233
 &0.2601&  0.8067&0.4858&0.3212& 0.3867&2.5201     \\
LongiFill  &  \textbf{0.3356}
& \textbf{0.1341}
& \textbf{0.2728} & \textbf{0.8219} &\textbf{0.5397} & \textbf{0.4178} &\textbf{0.471}   &  \textbf{2.6144}  \\ \hline
\end{tabular}

\caption{Comparative Study for evaluation between the Transformer and LongiFill Models: Metrics include NLG metrics such as BLEU (BL), METEOR (M), and ROUGE \(R_L\), CE metrics such as Accuracy (A), Precision (P), Recall (R), and F-1 score (F1), and our proposed evaluation method, Regressed GPT-4 (ReGPT-4).  }
\label{Comparative}
\end{table*}
\noindent \textbf{Sentence-Level Evaluation: Superior Correlation with Human Judgment}
In the process of evaluating the original-prediction pairs, each sentence was first assigned a corresponding sentence score. The agreement between these scores and human evaluations, analyzed on a sentence-by-sentence basis, was quantified using Cohen’s Kappa \cite{viera2005understanding} in Figure \ref{heatmap}, providing valuable insights.
\textbf{1. Agreement between the Ground Truth and Others:} Rater1, with expertise in radiology, exhibited the highest kappa scores with the ground truth, registering a kappa score of 0.956. Given their specialized training, this robust correlation is anticipated. Notably, the 5-shot version of the detailed GPT-4 model (with a score of 0.74) has achieved a performance remarkably close to that of Rater2 (0.76) and Rater3 (0.73).
\textbf{2. Effect of Different GPT Models: } GPT-4 displayed superior performance to GPT-3.5 across all conditions, implying improved capabilities in the newer model version. 
\textbf{3. Effect of Different Instructions:}
The effectiveness of detailed instructions is notably reflected in GPT-4's performance. For example, when using a Detailed GPT-4 approach with 5-shot learning, there was an improvement of 0.04 (from 0.70 to 0.74) compared to the simpler approach. This improvement can likely be attributed to the more detailed instructions providing clearer context, which enables GPT-4 to generate sentences that more closely align with the intended objective.
In contrast, the results are reversed when applying detailed instructions with GPT-3.5. This difference is likely due to GPT-3.5's relatively lower reasoning and learning capabilities compared to GPT-4. In the case of GPT-3.5, providing excessive context appears to hinder rather than enhance its performance, underscoring the nuanced differences between these two models in handling complex instructions. 
\textbf{4. Effect of Different Number of Templates:} In our experiments, we tested scenarios with a single template (1-shot learning) and with five templates (5-shot learning). Beyond the influence of instructions, we also noticed that the 5-shot learning method typically outperforms the 1-shot method for GPT-4. The 5-shot approach offers more examples to the model, aiding it in better comprehending the requirements of the task.

\noindent\textbf{Case Study}
In this specific case in Table \ref{case}, the BLEU, ROUGE, and METEOR metrics, which are common NLG metrics for natural language generation tasks, 
demonstrates the risks associated with depending solely on n-gram overlap between predicted and original sentences.
For example:
(1) The original report says ``the cardiac silhouette is enlarged'', while the generated sentences note ``moderate to severe cardiomegaly is re-demonstrated''. Here, both phrases indicate heart enlargement, but NLG metrics might fail to recognize this synonymy.
(2) The original report mentions ``there is again seen moderate <unk> pleural effusion which is stable'', whereas the generated sentences note ``small bilateral pleural effusions are noted''. These sentences are somewhat similar, but NLG metrics might not capture the difference in severity (``moderate'' vs ``small'').
Contrary to this, GPT-4 (In this case, we use detailed Gpt-4 (5-shot) to generate this evaluation result) showcases its prowess in discerning semantic similarities and differences that might elude conventional metrics. GPT-4's evaluations tend to align closely with human evaluations. This alignment suggests that GPT-4 captures not just general semantic understanding, but also clinical nuances pivotal to radiology and patient care.
To enhance comprehension, GPT-4 was also employed to generate explanations for the scores, reinforcing the criticality of precise interpretation in fields like radiology.  This is crucial in fields such as radiology, where the interpretation of findings can greatly influence patient care and results.

\noindent \textbf{Effect of Explanation}
\begin{table}[]
\centering
\small
\setlength\tabcolsep{0.9mm} 

\begin{tabular}{l|cc|cc|cc|cc}
\hline
          & \multicolumn{4}{c|}{GPT-3.5} & \multicolumn{4}{c}{GPT-4}  \\ \cline{2-9} 
          & \multicolumn{2}{c|}{w/ expl.} & \multicolumn{2}{c|}{w/o expl.} & \multicolumn{2}{c|}{w/ expl.} & \multicolumn{2}{c}{w/o expl.} \\ \hline
Kendall's & 0.1551 &  &-0.0030  &  & \textbf{0.4775}&  & -0.0232& \\
Pearsons  &0.2264&  &0.0130 &  &\textbf{0.6336}  &  &  0.0063& \\ \hline
\end{tabular}
\caption{Comparative analysis of Kendall's and Pearson's Correlation Coefficients: GPT-3.5 vs. GPT-4 in Agreement the ground truth's Overall Scores. ``w/ expl.'' refers to ``with explanation'' and ``w/o expl.'' refers to ``without explanation''.}
\label{explana}
\end{table}
We also conducted ablation experiments in LLMs directly predict the overall scores of reports without providing explanations. For these experiments, we employed \textit{Kendall's Tau} \cite{kendall1938new} and \textit{Pearson's Correlation Coefficient} \cite{pearson1895vii} to analyze the results.
Table \ref{explana} reveals that both models demonstrate improved performance when they include explanations, exhibiting stronger correlations with expert evaluations.
Notably, GPT-4 shows a significant enhancement in alignment with human expert evaluations when explanations are provided, compared to GPT-3.5. However, in scenarios lacking an explanatory framework, the near-zero correlation scores highlight that both GPT-4 and GPT-3.5 face challenges in aligning their assessments with those of human experts. This finding emphasizes the crucial role of a CoT or an explanatory framework in augmenting AI models with expert knowledge.
These results strongly support a collaborative approach between AI models and human experts. By combining the strengths of both, we can more effectively refine the evaluation of complex medical reports, leveraging the unique insights and capabilities of each.

\noindent\textbf{Comparative Study for evaluation between the Transformer and LongiFill Models}
To rigorously assess the efficacy of our newly proposed evaluation model, we conducted a comparative analysis between two distinct models:  the  advanced LongiFill model and its baseline transformer model in \cite{zhu2023utilizing} . The transformer is a simple model that uses image as input and output reports, trained on the same dataset as LongiFill. For a more comprehensive and robust evaluation, we randomly selected a  set of 300 examples generated from each model.
Considering the results in Table \ref{Comparative}, the LongiFill model consistently outperformed the baseline, particularly in established metrics such as BLEU and ROUGE. This superiority was not merely restricted to traditional metrics. Impressively, when assessed using our novel evaluation model, the LongiFill model's scores witnessed an improvement. Such results serve as a testament to the reliability  of our proposed evaluation framework.


\section{Conclusion}
In this paper, we introduced a novel method for evaluating AI-generated radiology reports, by leveraging the expertise of professional radiologists and the capabilities of large language models. Our method demonstrated superior performance over traditional metrics and a high correlation with human evaluations. Furthermore, our method is explainable, providing valuable insights that can be used to improve the AI models generating the reports. We believe our work contributes to the ongoing advancements in the field of AI and healthcare, paving the way for more reliable, accurate, and trustworthy AI applications in medical report evaluation.

\section{Limitations}
Our study provides valuable insights but also faces certain limitations. Our methodology, specifically tailored for chest X-ray report evaluation, utilizes the most extensive dataset currently available in this field. This approach, owing to the standardization in radiological practices, holds promise for application to various chest X-ray datasets globally. Furthermore, the consistency in medical terminology indicates the potential applicability of our method to other types of imaging reports, such as those from CT scans. However, further testing on a broader range of datasets is essential to confirm this potential. The generalizability of our approach across different types of radiology reports or other medical fields is not yet fully determined, necessitating additional research to evaluate its transferability and effectiveness in contexts beyond chest X-rays.
\section{Ethical Statement}
Our experiments strictly followed HIPAA compliance through the Azure OpenAI Service and adhered to the PhysioNet Data Use Agreement, ensuring the confidentiality of MIMIC data. We accessed GPT models via Azure's OpenAI service, taking necessary steps to keep our data private and unreviewed by Microsoft, as per our agreement.
\section{Acknowledgements}
   This research was supported by the Intramural Research Program of the National Library of Medicine and Clinical Center at the NIH.

\bibliographystyle{IEEEtran}
\bibliography{anthology,custom}

\appendix

 \section{Appendix}
 Table \ref{Prompt} presents both the Simplistic and Detailed versions of task instructions that guided LLMs through the evaluation process using in-context learning. Table \ref{Templete} provides a template used during the evaluation, while Table \ref{re-write} showcases the instructions and template for the optimized process. Table \ref{regress} provides Kendall's Tau values and corresponding p-values for various regressed models. Random Forest performs best. Also, the various regression models exhibit comparable performance, demonstrating the robustness of our method.
 
\begin{table*}[!h] \small
\centering

\begin{tabularx}{\textwidth}{c|X}
\hline
\multicolumn{2}{c}{Instructions}
\\
\hline
Simplistic & Your task is to evaluate the prediction sentence by sentence, comparing it with the original report. You should also have one overall score from 0-5 for prediction sentences with respect to the original report based on your subjective impression. And I only need the csv format for the output.\\
\hline
Detailed & In this task, you are required to evaluate pairs of medical reports: an original report and a predicted report. The original report is the benchmark, containing confirmed and accurate information about a patient's condition derived from radiological studies. The predicted report, generated by another model, requires assessment for its degree of accuracy. Each report includes several points or observations about the patient's health, designated by identifiers such as ``a'' , ``b'' , ``c''  for the original report and ``A'' , ``B'' , ``C''  for the predicted report. Your objective is to appraise the accuracy of each point in the predicted report concerning the equivalent point in the original report. Please adhere to the following scoring guidelines: 
""

Score of 1:

The predicted statement matches the original in meaning and details.
Even if words are different, the message is the same.
No information from the original is missing or contradicted.

Score of 0.5:

The predicted statement somewhat resembles the original statement but not entirely.
Some elements align, but others might differ or be missing. The core idea might be there, but not all details are accurate.

Score of -1:

The predicted statement goes against or is the opposite of the original  statement.
The information presented directly contradicts what's in the original report.

Score of 0:

The predicted statement doesn't have a matching original  statement to compare with.


Beyond the detailed scoring for each prediction statement, you are also required to give an overall score ranging from 0 to 5 for all the prediction sentences with respect to the original report. Your task is not only to find errors or omissions, but also to assess the clinical relevance, accuracy, and potential harm of the AI's outputs. Please bear in mind that incorrect information in a clinical setting could be potentially more damaging than missing information. It's not just about comparing the AI and human reports line by line. Please consider the whole report and its coherence, the context of the patient's condition, and the potential impact of the AI statements on clinical decisions. Remember, you're evaluating these reports as if they were to be used in a real-world clinical scenario. After your evaluation, please provide an overall impression score of the AI's report from 0 to 5, where 0 is 'poor' and 5 is 'excellent'. Keep in mind that these reports could be used by other healthcare professionals (such as physicians, surgeons, and nurses) and mistakes or unclear information could lead to wrong treatment decisions. Your insights as a clinician are crucial in refining this AI tool and improving patient outcomes.

\\
\hline
\end{tabularx}
\caption{ Simplistic and Detailed Instructions provided by radiologists for the LLMs to evaluate medical report predictions.}
\label{Prompt}
\end{table*}

\begin{table*}[h!]\small
\begin{tabularx}{\textwidth}{c|X}
\hline
\multicolumn{2}{c}{A Template}\\
\hline
Input &Original ``a. In comparison with a series of images from and there has been progressive decrease in the pleural fluid in the left hemithorax though some persists b. Elevation of the hemidiaphragm with mild shift of the mediastinum to the left is consistent with previous surgery c. The right lung is clear and there is no vascular congestion.'' \\

&Prediction  ``A. PA and lateral views of the chest were reviewed and compared to the prior studies B. Elevation of the left hemidiaphragm is unchanged since C. The right lung is clear D. There is no pulmonary edema vascular congestion pleural effusion or pneumothorax E. The cardiac and mediastinal contours are normal F. There are no concerning osseous or soft tissue lesions.''\\
\hline
Output& \noindent Original ,a,in comparison with a series of images from and there has been progressive decrease in the pleural fluid in the left hemithorax though some persists,D,0.5,

\noindent Original,b,elevation of the hemidiaphragm with mild shift of the mediastinum to the left is consistent with previous surgery,B,1,

\noindent Original,c,the right lung is clear and there is no vascular congestion .,``C,D'',1,

\noindent Original,A,pa and lateral views of the chest were reviewed and compared to the prior studies,a,0.5,

\noindent Original,B,elevation of the left hemidiaphragm is unchanged since,b,1,

\noindent Original,C,the right lung is clear,c,1,

\noindent Original,D,there is no pulmonary edema vascular congestion pleural effusion or pneumothorax,``a,c'',0.5,

\noindent Original,E,the cardiac and mediastinal contours are normal,0,

\noindent Original,F,there are no concerning osseous or soft tissue lesions .,0,

\noindent ,,-,,,3.5/5, ``The AI-generated report is quite consistent with the original report. Both reports note the elevation of the left hemidiaphragm and the clear state of the right lung. However, there are minor differences. The original report mentions a progressive decrease in the pleural fluid in the left hemithorax, which the AI does not explicitly point out. Instead, the AI report states there is no pleural effusion, which might imply the resolution of previous fluid. Additionally, the AI report adds observations on cardiac and mediastinal contours and the absence of concerning osseous or soft tissue lesions, which are not mentioned in the original report.''
      
\\
\hline
\end{tabularx}
\caption{Template for Evaluating Medical Reports: This table showcases the evaluation process where reports are input and the results are output in a CSV format. 
}
\label{Templete}
\end{table*}

\begin{table*}[!ht]\small
\begin{tabularx}{\textwidth}{c|X}
\hline
Instruction &
You are provided with pairs of chest X-ray reports and their corresponding reviews. Based on the observations and corrections mentioned in the reviews, your task is to create a revised report for each original report. Ensure that the revised report accurately reflects the observations from the review, including additions, omissions, and corrections.
For each report:
Begin with a label like ``Revised Report,'',Follow it with the revised content.
Your goal is to ensure that the revised report is clear, concise, and written in a clinically appropriate manner using correct medical terminologies.
\\
\hline
Template &

Report,``pa and lateral chest views were obtained with patient in upright position . analysis is performed in direct comparison with the next preceding similar study of . the heart size remains normal . no configurational abnormality is identified . thoracic aorta unremarkable . the pulmonary vasculature is not congested . no signs of acute or chronic parenchymal infiltrates are present and the lateral and posterior pleural sinuses are free . no pneumothorax in apical area . skeletal structures of the thorax grossly unremarkable .''

Review,``The AI report correctly notes the absence of pleural effusions, pneumothorax, and normal cardiac and mediastinal contours. However, it does not mention the right upper lobe consolidation with air bronchograms, the new focal tubular lucency within the opacity, and the progressed opacity in the right lower lobe. Furthermore, the AI report fails to capture the original report's note on the mild thickening of the left major fissure. Finally, the AI report adds unnecessary details not mentioned in the original report, such as the unremarkable thoracic aorta and the absence of acute or chronic parenchymal infiltrates.''

Refined Report,``pa and lateral chest radiographs were obtained . a right upper lobe consolidation with air bronchograms is similar to . focal tubular lucency within the opacity is new and may reflect cavitation dilated airways or spared lung parenchyma . opacity in the right lower lobe has progressed since the prior study . there is no effusion or pneumothorax . cardiac and mediastinal contours are normal . there is mild thickening of the left major fissure . ''
      
\\
\hline
\end{tabularx}
\caption{Instruction and Example Template for the refined process. This table provides guidelines on how to integrate observations and corrections from reviews into the AI generated reports to produce clinically accurate refined reports. The template illustrates the refined process with an example.}
\label{re-write}
\end{table*}
\begin{table*}[!ht]\small
\centering
\begin{tabular}{lrr}
\toprule
{} & Kendall's Tau Value & P-Value \\

Decision Tree & 0.5659 & 1.5435e-13 \\
Support Vector Machine & 0.6175 & 1.0276e-17 \\
K-Nearest Neighbors & 0.6204 & 2.4497e-17 \\
Neural Network & 0.6216&6.0731e-18 \\
Gradient Boosting & 0.6231 & 9.8870e-19 \\
Random Forest & 0.6372 & 9.3613e-19 \\
\bottomrule

\end{tabular}

\caption{Kendall's Tau values and corresponding p-values for various regressed models.}
\label{regress}
\end{table*}

\end{document}